# Phasic dopamine release identification using ensemble of AlexNet

Luca Patarnello[1], Marco Celin[2], Loris Nanni[3;]

[1] DEI, University of Padua, viale Gradenigo 6, Padua, Italy.
Email: luca.patarnello@studenti.unipd.it
[2] DEI, University of Padua, viale Gradenigo 6, Padua, Italy.
Email: marco.celin.1@studenti.unipd.it
[3] DEI, University of Padua, viale Gradenigo 6, Padua, Italy.
Email: loris.nanni@unipd.it
*Corresponding Author

## Abstract

Dopamine (DA) is an organic chemical that influences several parts of behaviour and physical functions. It also plays significant roles in pathologies such as schizophrenia, Parkinson's disease and drug addiction. Fast-scan cyclic voltammetry (FSCV) is a technique used for in-vivo phasic dopamine release measurements. The analysis of such measurements, though, requires notable effort. In this paper, we present the use of convolutional neural networks (CNNs) for the identification of phasic dopamine releases. In particular, an ensemble of classifiers combined by the sum rule fusion of several AlexNet networks, trained using images from the DA dataset, saliency detection methods and the YOLOv2 object detector. The ensemble we produced was able to outperform previous known methods both in accuracy and simplicity in its application. The code developed for this work will be available at https://github.com/LorisNanni.

## 1. Introduction

The neurotransmitter dopamine (DA) is deeply involved in the action-selection dedicated part of the brain system, acting as main regulator in core body and brain functions such as movement, learning and memory [1]. Dopamine release can occur in two different ways [2]. Phasic DA release is distinguished by a high concentration of dopamine being rapidly released into the system. Tonic DA release happens in small concentrations distributed over a longer period of time. Abnormal alterations in dopamine levels are associated to neurological disorders, including Parkinson's disease, schizophrenia and substance dependence [3]. Although dopamine has been subject of numerous neuroscientific researches through the years, its precise behaviour and effect is only partially known. This is why, in order to gain better knowledge on the matter, it is crucial to elaborate a reliable method to accurately measure synaptic DA release.

A method used to perform such reading is fast-scan cyclic voltammetry (FSCV). FSCV is an electroanalytical technique used to detect neurotransmitters concentrations such as dopamine [4]. This is accomplished by measuring the variation of current in the electrodes, which is proportional to the DA level. FSCV data is stored in a numerical matrix, which can be processed into images represented by the applied potential on the y-axis and the cycle (time) on the x-axis, with the current being represented by the pixel intensity. Although fast-scan cyclic voltammetry can be considered to be more appropriate than most techniques for the measure of phasic dopamine release, it produces a great amount of data to be analyzed. When the task of analyzing such data is to be fulfilled manually, even for an expert would result in being a long and redundant process. Therefore, research was encouraged for the implementation of automated solutions to apply in the context of phasic DA release identification [1].

The employment of deep learning for image classification has been producing new and highly accurate ways to automatize image analyzing and object recognition. Matsushita et al. [1] proposed a combined CNN approach, which consisted in a first classification using full images, followed by a second classification using extracted patches. The model, trained and tested with a 10-fold division testing protocol, obtained an accuracy of 97.35%.

The approach we propose consists in developing an ensemble of AlexNet, based on the methods described in [1] and saliency detection methods, which are algorithms that display the most relevant regions of an image and object detectors. Our experiments proved to be successful, producing a model with an accuracy of 98.75%, following a 10-fold division testing protocol, which is the fairest, since it makes sure that all the recordings from a single experiment are contained in the same fold and are not present in both training and testing at the same time. The experiments were performed using images from the latest dataset introduced in [1], which is, to date, the only known public dataset of FSCV images. The dataset is available at https://web.inf.ufpr.br/vri/databases/phasicdopaminerelease/.

## 2. Deep Learning & CNNs

Deep learning is a class of machine learning algorithms based on artificial neural networks and feature learning [5] [6]. A convolutional neural network (CNN) is a prominent deep learning tool that has been proved to be remarkably effective when implemented in imaging and computer vision applications such as medical image analysis. The main advantage of a CNN compared to other image classification algorithms, is that it automatically extracts the most important features, therefore requiring significantly less pre-processing. In this paper we use the CNN named AlexNet.

### 2.1. AlexNet

AlexNet is a noteworthy CNN model, presented by Krizhevsky et al. in 2012 [7]. In the same year, it won the ImageNet large scale visual recognition challenge [8], significantly outperforming every other image classification algorithm competing. The architecture has a depth of eight layers: five convolutional layers, with some of them followed by a max-pooling layer, three fully-connected layers. ReLU nonlinearity is applied to each layer as activation function. Finally, classification is performed by a softmax layer, fed by the output of the last fully-connected layer.

## 3. Dataset & Protocol

### 3.1. DA Dataset

The dataset utilized for our work is the latest introduced by Matsushita et al. [1]. FSCV datas are used to generate images representing plots of 20 seconds long experimental recordings. These recording were performed by the Laboratory of Central Nervous System of the Federal University of Parana (UFPR) at Curitiba, Brazil and from D. Robinson's Laboratory of the University of North Carolina (UNC) at Chapel Hill, United States of America. Different animals and FSCV setups were used by the two laboratories, producing slightly different looking plots. A total of 30 experimental recordings were performed. The UFPR laboratory used 29 male Swiss mice, one for each experiment and the UNC laboratory used 6 male Sprague Dawley rats, all in the same experiment.

For each image, columns were selected from the beginning (0.5s), middle (10s) and end (19.5s) of the image, then their values are subtracted from the other columns, generating 3 different background (A, B, C) and thus, 3 different images. A standard false color palette used by FSCV analysis softwares was then applied to better display transitions. For each background, there are a total of 1005 evoked dopamine release images and 1005 images without dopamine release. Each image is identified by its experiment, which is also distinctive for the animal used in it, except for the experiment "RIX2" performed by the UNC laboratory, since it contains all the mixed readings obtained from the different animals they used.

All experiments were performed in accordance with the NIH Guide for the Care and Use of Laboratory Animals with procedures approved by the Institutional Animal Care and Use Committee of the University of North Carolina, and the Institutional Ethics Committee for Animal Experimentation of the Federal University of Parana (Protocol 638).

### 3.2. Testing Protocol

All the experiments were run on pre-trained AlexNet networks using transfer learning. Since the network is initially configured on 1000 classes, we fine-tuned the last three layers to adapt them to our problem. Also, input images are resized to the network established input size of 227x227 pixels.
For the training of the networks, a 10-fold cross validation is implemented. Images from the same experimental recording are placed in the same fold, in order to prevent the presence of samples from the same animal in both training and testing.

The hyperparameters set for the networks are the following:
- miniBatchSize = 30
- learningRate = 1e-4
- optimizer = 'sgdm'
- maxEpochs = 20

They are a standard also used in other works [9]. No episodes of overfitting were recorded using the dataset.

## 4. Proposed Methods

### 4.1. Matsushita et al. methods

With the availability of a new and larger dataset, Matsushita et al. were presented with the opportunity of new approaches [10][1]. We branched those methods into two categories: Global methods and Patch methods.

### 4.1.1. Global methods: Original images and zoning variations

The Global category consists of 3 methods. The original images method uses images in their entirety. The first zoning method consists in only the section of the image from pixel 320 to pixel 520 of the y-axis. This is because it was visually noticed that phasic DA release episodes commonly manifested in that region, hence the name "common dopamine release region". The image size for this method is 875x200 pixels. The second zoning method consists in using, in addition to the common dopamine release region, the first 90 pixels of the y-axis of the image, which in some releases contains visual information that could be useful for feature extraction. This region is called "concatenated zones". The image size for this method is 875x290 pixels. A graphic description of the global methods is shown in the figure 1.

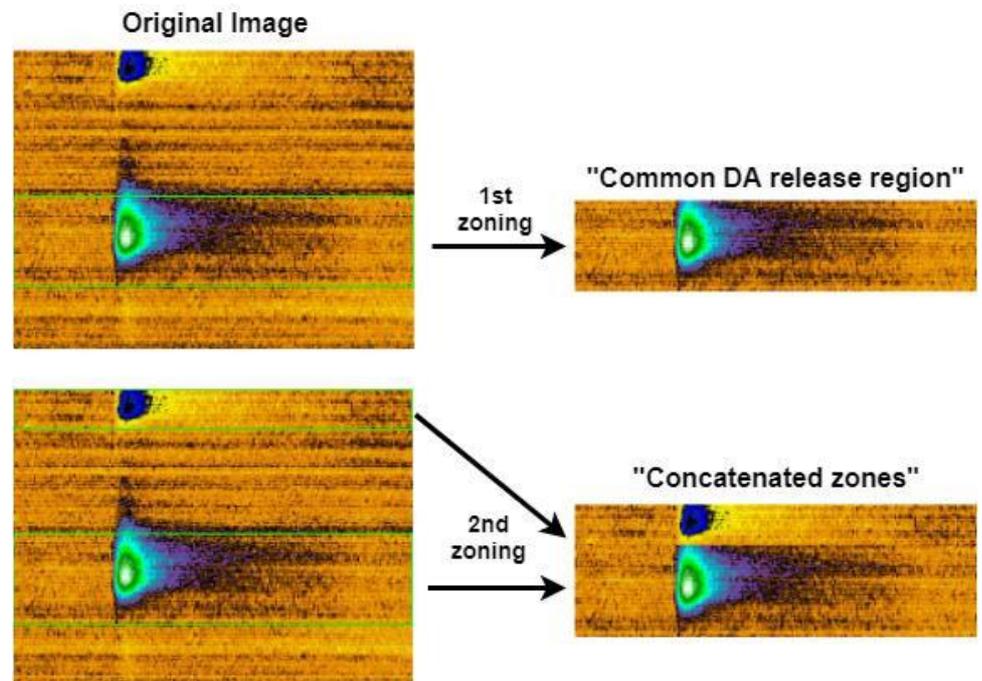

Figure 1. Original Image and Zoning variations

### 4.1.2. Patch extraction methods

Patches of size 200x200 are manually and automatically extracted from the common release region. Manual patches are extracted using the release peak position label included in the dataset, setting it as the center of the image. Automatic patches are extracted applying a sliding window of size 200x200, and window slide of 135 pixel over the common release region. The slide size is justified by being a divider of the width size of the original image, therefore, there are no pixels leftover and six patches extracted per image are provided.

In addition to the size 200x200 patches, in this paper we also propose the extraction of size 290x290 patches from the concatenated zones. While the approach for the manual patches remains unaltered, since we can simply change the width and height of the image to 290 pixels, the sliding process still uses a 135 pixel slide, hence, automatic extraction provides patches of size 200x290, to which a 0-padding is applied to make them of size 290x290. Patches examples are shown in figure 2.

Special attention shall be given to the fact that, when this method is implemented, patches manually extracted are used exclusively for training, while patches automatically extracted exclusively for testing. Automatic patches are dependent from the image they have been extracted from, consequently, given an image from which patches have been automatically extracted, its score is determined by applying the max rule over the patches scores, where the score of the patch, with the highest similarity to the DA release class, will be picked.

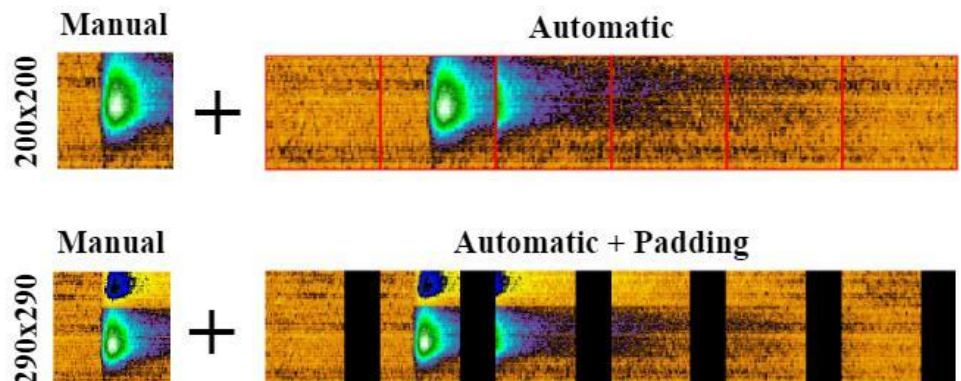

Figure 2. Patches variations

### 4.2. Saliency detection methods

Saliency detection is a particular category of computer vision algorithms, which goal is to simulate the visual attention of Human Visual System (HVS), in order to retrieve the most important features in what could be an over-informative environment [11]. Algorithms of this kind can be of two types: the bottom-up approach, which is stimulus-driven and aims to replicate the pre-attentive stage of the HVS and top-down approach, which is goal-driven and resembles the behaviour of the attentive stage of the HVS [12].

In this paper, we used five different saliency detection approaches: SimpSal, GBVS, Co-Saliency, SPE and Wavelet Transform. A sample of their saliency mask is displayed in figure 4. Every saliency method can be used to create a binary mask, by setting to 1 all the pixels of the produced saliency mask, whose values are above a fixable threshold, and setting to 0 all the others [13]. We use each saliency method on every image from the dataset. When a saliency method is applied to an image, three different output images are returned: foreground (FG), region of interest (ROI) and foreground region of interest (FG-ROI). We demonstrate the saliency detection process in figure 3. The FG image is obtained by multiplying the original image by its binary mask. The foreground, which is composed by the non-salient pixels of the image, is set to black by this operation. To obtain the ROI image, in addition to the same operation use to obtain the FG image are applied, colums and rows that are almost or completely black are cut from the final image.

The FG-ROI image is obtained in the same way as ROI, with the except of the pixels all maintaining their original value, regardless of the values of the correspondent binary mask. The pseudocode, of the functions just described, is shown in algorithm 1, algorithm 2 and algorithm 3.

```
Function foreground
    Input: Original image O, binary mask B
    Output: FG image
    for x in row do
        for y in column do
            FG(x; y) = O(x; y)*B(x; y);
        end
    end
    return FG;
end
```
Algorithm 1: FG

```
Function fg roi
    Input: Original image O, binary mask B, threshold TH
    Output: FG ROI image
    separate color channels from O;
    for each column in B do
        sum column values;
        if sum < TH then
            add column to columns to be removed;
        end
    end
    remove columns to be removed from color channels;
    for each row in B do
        sum row values;
        if sum < TH then
            add row to rows to be removed;
        end
    end
    remove rows to be removed from color channels;
    FG ROI = combined color channels;
    return FG ROI;
end
```
Algorithm 2: FG ROI

```
Function roi
    Input: Original image O, binary mask B, threshold TH
    Output: ROI image
    FG = foreground(O; B);
    ROI = fg roi(FG; B; TH);
    return ROI;
end
```
Algorithm 3: ROI

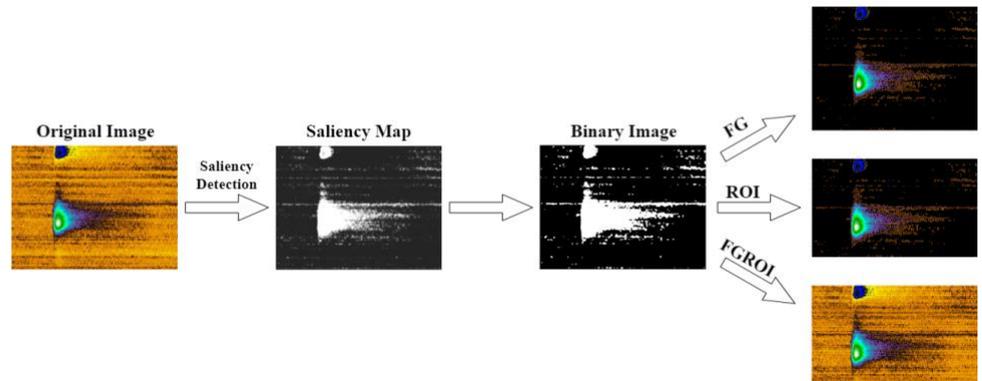

Figure 3. Saliency detection pipeline

### 4.2.1. SimpSal

This method is the Matlab implementation that is most faithful to the model of Itti's et al. [14]. This specific saliency map follows a bottom-up approach. It aims to represent the conspicuity at every location in the visual field by a scalar quantity and to guide the selection of attended locations, based on the spatial distribution of saliency.

In the proposed model, the first step consists in computing a total of 42 early visual feature maps, 6 for intensity, 12 for color and 24 for orientation, 6 for each preferred orientation degree (0°, 45°, 90°, 135°). Across-scale addition, applied to reduce each map to scale four and point-by-point addition, is then used to combine feature maps into three "conspicuity maps," one for intensity, one for color and one for orientation. For the orientation map, additional operations are computed:
for each orientation degree, the six corresponding feature maps are combined to create four intermediary maps, which are then combined into a single orientation conspicuity map. Finally, the result obtained after the three conspicuity maps are normalized and summed, constitutes the final input to the saliency map (SM). The SM is modeled as a 2D layer of leaky integrate-and-fire neurons at scale four. These model neurons consist of a single capacitance which integrates the charge delivered by synaptic input, of a leakage conductance, and of a voltage threshold. When the threshold is reached, a prototypical spike is emitted and the capacitive charge is ruled out to 0. At any given point in time, the focus of attention, which is a disk with fixed radius, is directed by the maximum of the SM, which defines the most salient area of the image at that moment.

### 4.2.2. GBVS

Graph-Based Visual Saliency (GBVS) [15] is a bottom-up visual saliency model that primarily relies on two steps: first creating activation maps, using feature channels, to create a dissimilarity measure on the pixels of the image and then normalizing them. The goal set when forming the activation map is to do that is such a way that, intuitively, high values of activation (A) correspond to points in the feature map (M) that are in some way unusual in its neighborhood. To fulfill the first step, an organic Markovian approach is proposed.

Then, a Markov chain is defined by normalizing the weights of the outbound edges to 1. The equilibrium distribution of this chain, would naturally accumulate mass at nodes that have high dissimilarity with their surrounding nodes, since transitions into such subgraphs is likely, and unlikely if nodes have similar M values. To acquire the saliency map, the last step of normalizing the Markov chain obtained is the same way it was before, with the difference that this time the weights are assigned proportionally to the activation values of the activation map, hence, the mass of the equilibrium distribution of this chain will accumulate at points with higher activation values.

### 4.2.3. Cluster-base saliency detection (COS)

Co-saliency algorthms are used to discover the common saliency in a group of similar images. Huazhu et al. [16] presented the cluster-based algorithm for co-saliency detection, which differentiates itself from other co-saliency algorithm by removing the constraint of similarity between the images in the group [16]. An additional constraint is found in implementing repetitiveness property, to discover common salient object on the multiple images. The proposed model first employ two clustering layers, one works on single images, grouping the pixels on each of them and the other to associates the pixels on all images. The cluster-level saliency is then measured by computing three saliency cues:

- Contrast Cue: represents the visual feature uniqueness on the single or multiple images.
- Spatial Cue: Simulates the human visual system in order to remove the salient background detected by the contrast cue and highlights the region at the center of the cluster. Like contrast cue, it can be used for both single and multiple images.
- Corresponding cue: measures how frequently the cluster recurs on the multiple images
  and how it is distributed.

Each cue is finally normalized before being integrated, using a multiplication operation, that will return the co-saliency map.

### 4.2.4. Spectral residual (SPE)

The approach to saliency detection using Spectral Residual, presented by Hou and Zhang in [17], consists in an algorithm that analyzes the log-spectrum of an input image in spectral domain, extracts the spectral residual and constructs in spatial domain the corresponding saliency map. The statistic average of a signal, defined as its frequency content, is called its spectrum. The scale invariance property states that the amplitude of the averaged Fourier spectrum is a feature that does not change in the object, if the scale changes.

The log spectrum representation of the image is adopted to analyze its scale invariance. In particular, the attention is directed towards the smooth curves, since its believed that they contain information about statistical singularities, which are indicators of anomalous region where proto-objects are present. The log spectrum is computed from the input image, down-sampled to a height (or width) of 64 pixels.

### 4.2.5. Wavelet Transform

Wavelet transform (WT) has been subject of numerous researches for visual attention modeling, given its ability to provide multi-scale spatial and frequency analysis simultaneously. Imamogluˇ et al. [11] introduced a saliency detection model based on high-pass coefficients of the wavelet decomposition. By applying Inverse WT on different composition levels, this model is able to create more detailed feature maps, compared to others with similar approach. This gives the advantage to use different bandwidths when looking for irregularities. The model initially creates two saliency maps, one for local saliency and one for global saliency. The reasons for two different saliency maps is to perform separate normalization of each feature map and to make sure that both are properly taken into consideration. The global saliency is computed using a Gaussian probability density function in multi-dimensional space. The local saliency is created by taking in consideration the maximum value between pixels of the different feature channels.

The local and global saliency maps, are then combined to obtain the final saliency map. A last enhancement process is fulfilled, in order to improve the focus of attention towards the detected salient areas.

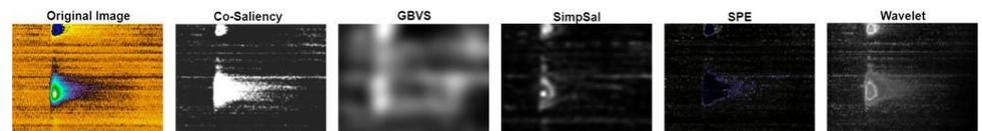

Figure 4. Saliency Map Examples

### 4.3. Object detector: YOLOv2

Object detection is a development of deep learning which goal is to locate existing objects within an image, classify them and label them with rectangular bounding boxes and the respective confidence of existence score assigned by the detector [18]. There are two types of frameworks that categorize object detection. One initially generates region proposals and then classifies each of them into different object categories. The other assesses object detection as a regression/classification problem, in which objects are directly categorized and located in a combined approach.

You Only Look Once (YOLO) is an object detector proposed by Redmon et al. that adopts the regression/classification based framework [19]. YOLO initially divides the input image into an SxS grid and each grid cell is responsible for predicting objects centered in it. Each grid predicts a fixed number B of objects and their confidence scores. The confidence score shows how confident is the detector of the presence of the object in the box and how accurate it is. At the same time, C conditional class probabilities are predicted in each grid cell. This is the probability that the object detected belongs to a specific class. At test, the multiplication of confidence score and class probability returns the class-specific confidence score for each box. YOLOv2 is the follow-up and improved version of YOLO [20]. Unlike YOLO, YOLOv2 uses anchor boxes to predict offsets instead of directly predicting the coordinates of the bounding boxes. To improve accuracy, the new version also adopts Batch normalization, high-resolution classifier and dimension cluster.

In this paper, we utilize a matlab implementation for the YOLOv2 detector, utilizing the release intervals labels included in the dataset to create the bounding boxes for training. Up until now we classified images in those having a phasic DA release and those that have not. We cannot face the problem in the same way with YOLOv2, since we have to detect objects. For this reason, the detector will have to detect object from only one class, thus, it will be trained only on images with release. Testing is performed with images from both classes. If an object is detected, it is classified as a release if its confidence score is higher than the fixed threshold of 50%. When multiple bounding boxes are assigned, we consider only the one with the highest confidence score. If the confidence score of the bounding box delimiting the object is lower than said threshold or no object is detected, the image is classified as having no release.

## 5. Proposed ensemble

Each of the methods described are further applied on the images without release included in the DA dataset. This maintains class balance in the datasets variations obtained. 20 datasets are therefore produced: three by the global methods, two by the patch methods and three by each of the five saliency detection methods, as shown in figure 5. By using the same procedure on each of the three background variations, a total of 60 datasets is obtained.

We trained AlexNet on each one of those datasets and YOLOv2 on the original images leading us to have 60 different CNNs and 3 different YOLOv2 detectors with the same architecture trained on different datasets. In [9], Nanni et al. show the effectiveness of fusing the scores of different CNNs with the same architecture trained on augmented data. The scores of all the CNNs and the detectors are therefore combined using the sum rule, which consists in summing the output of the softmax layers of the architectures to obtain a new score vector. The highest score dictates the prediction of the ensemble.

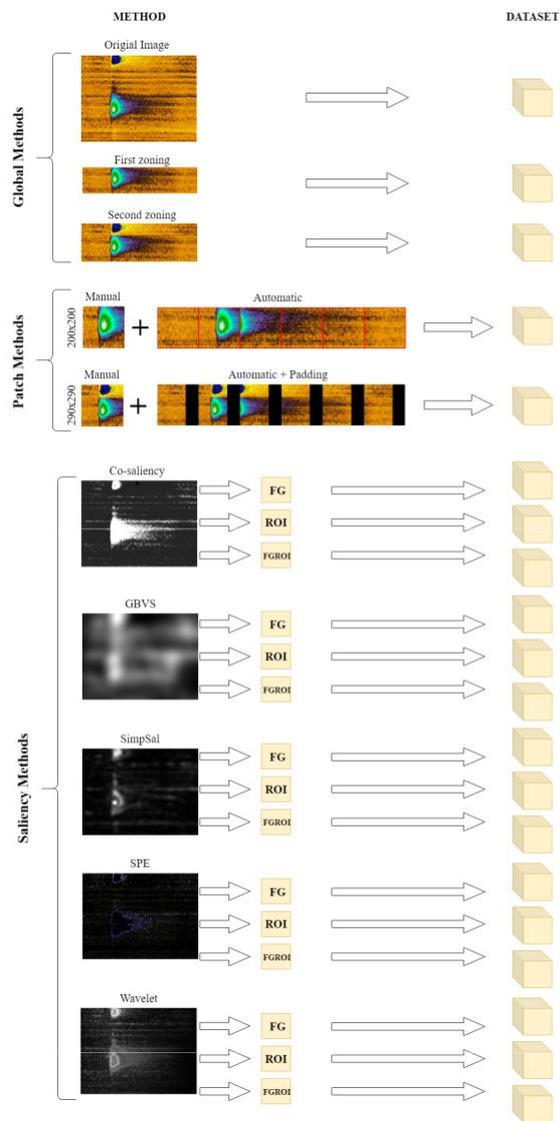

Figure 5. Dataset variations for a single background variation

## 6. Experimental results

The results of the different ensembles, obtained from the fusion using sum rule of the proposed methods, trained on different backgrounds, are shown in table 6. Along with the accuracy of the models, other evaluation metrics are also included [21]. Sensitivity and specificity indicate the true positive and negative rate, respectively. AUC, often used for binary classification, indicates the capability of the model of distinguishing between classes. F1 score is the harmonic mean between precision and sensitivity (sometimes called recall). First, results from the single global methods trained on background A are displayed, where O is the original images method, 1 is the first zoning method, 2 is the second zoning method and Global is the fusion of all the global methods.

The CNNs trained with the original images and first zoning methods perform slightly better than the second zoning method, but we can already see improvements with the simple ensemble of the three methods. With the Global fusion of all the background variations, we see the first significant improvement, with the accuracy of the model reaching a value of 98.38%. Although, as we can clearly see on table 6, ensembles of classifiers obtained with patch methods and the YOLOv2 detector do not perform as well as the one obtained with global methods, when fused with the sum rule, they still have role in enhancing the accuracy of the model. The "Scores Fused" index of table 6 indicates the number of CNN/Object detector scores fused with sum rule. It is worth saying that, even considering the high number of fused elements in the ensemble, AlexNet performs considerably fast. For a single image, the elapsed time is 0.010479 seconds, testing on a NVIDIA GeForce GTX 1080 GPU, which add up to a very affordable execution time of 0.66 seconds for all 63 images. The classifier obtained with the fusion of all the methods proposed is called AllMethods. This is the final classifier obtained after running all the experiments and has an accuracy of 98.75%, which is 1.4% more accurate than the combined CNNs approach presented by Matsushita et al. [1], following the same 10-fold division testing protocol. It is also 0.44% more accurate than their best model, which is based on a 3-fold division protocol and does not have the same level of fairness as our protocol, since it uses images from the same experiment in both training set and test set.

| Background | Method | Scores Fused | Accuracy | AUC | F1 | Sensitivity | Specificity |
|---|---|---|---|---|---|---|---|
| A | O | 1 | 96.67 | 98.95 | 96.19 | 95.76 | 3.41 |
| A | 1 | 1 | 97.68 | 99.37 | 97.55 | 97.90 | 2.77 |
| A | 2 | 1 | 97.20 | 99.45 | 97.08 | 96.65 | 2.51 |
| A | Global | 3 | 97.85 | 99.50 | 97.62 | 97.33 | 2.10 |
| A+B+C | Global | 9 | 98.38 | 99.78 | 98.21 | 98.31 | 1.89 |
| A+B+C | Patch | 6 | 94.61 | 98.76 | 94.31 | 95.42 | 6.61 |
| A+B+C | YOLOv2 | 3 | 94.45 | 98.26 | 95.31 | 98.82 | 7.45 |
| A+B+C | Global+Patch | 15 | 98.51 | 99.76 | 98.30 | 98.89 | 2.26 |
| A+B+C | Global+Patch+Saliency | 60 | 98.74 | 99.84 | 98.65 | 99.00 | **1.68** |
| A+B+C | **AllMethods** | **63** | **98.75** | **99.87** | **98.65** | **99.29** | 1.96 |

Figure 6. Results Table

## 7. Conclusions

The purpose of this paper was to create an ensemble of classifiers for the identification of phasic dopamine releases. After employing different methods of image preprocessing, saliency detection and object detection, we obtained a final ensemble of 60 AlexNet networks and 3 YOLOv2 detectors. The ensemble managed to outperform previous approaches to this problem, using the same dataset. Along with the accuracy, ease-of-use was also improved, since only one classification task is performed, instead of the two classifications, using original images and patches, performed in [1]. Every method proved to be useful for the overall improvement of the ensemble performance, even if the accuracy of some was a few percentage point lower than others. Additional methods to isolate the ROI of the image will be the focus of future works, in order to provide new training procedures to further improve our model performances.

## Conflicts of Interest

There is no conflict of interest to declare.